%% file: 2024-icra-rueckin.tex
\documentclass[letterpaper, 10pt, journal, twoside]{style/IEEEtrans}     

\makeatletter
\let\NAT@parse\undefined
\makeatother

\usepackage{dblfloatfix}

\usepackage[numbers,sectionbib,sort&compress]{natbib}
\usepackage{bm}
\usepackage{gensymb}
\usepackage{graphicx}
\usepackage{amsmath}
\usepackage{amssymb}
\usepackage{subcaption}
\usepackage{amsfonts}
\usepackage{siunitx}
\usepackage{makecell}
\usepackage{multirow}
\usepackage{booktabs}
\usepackage{upgreek}
\usepackage[font=small]{caption}
\usepackage[export]{adjustbox}
\usepackage{tikz}
\usepackage{tabularx}
\usepackage{svg}
\usepackage{sidecap} \sidecaptionvpos{figure}{c}
\usepackage{hyperref}
\hypersetup{colorlinks, breaklinks, linkcolor=[rgb]{0.5,0.,0.}, citecolor=[rgb]{0.000,0.427,0.173}, urlcolor=[rgb]{0.031,0.318,0.612}}

\usepackage[nameinlink, capitalize]{cleveref}
\usepackage{color}
\definecolor{CommentPink}{rgb}{1,0.2,0.5}
\definecolor{CommentBlue}{rgb}{0,0,1}
\definecolor{CommentGreen}{rgb}{0,1,0}

\Crefname{section}{Sec.}{Sec.}
\Crefname{equation}{Eq.}{Eq.}

\DeclareMathOperator*{\argmax}{argmax}

\usepackage[printonlyused,withpage,nolist,nohyperlinks]{acronym}
\input{additionals/common_acronyms.tex}

\IEEEoverridecommandlockouts                           

\title{Semi-Supervised Active Learning for Semantic Segmentation in Unknown Environments\\ Using Informative Path Planning}
\author{\hspace{5mm} Julius R\"{u}ckin \hfill Federico Magistri \hfill Cyrill Stachniss \hfill Marija Popovi\'{c} \hspace{5mm}
\thanks{Manuscript received: Dec 07, 2023; Accepted: Jan 22, 2024. This paper was recommended for publication by Editor Aniket Bera upon evaluation of the Associate Editor and Reviewers’ comments.}
\thanks{All authors are with the University of Bonn, Cluster of Excellence PhenoRob, Institute of Geodesy and Geoinformation. Cyrill Stachniss is also with the University of Oxford and Lamarr Institute for Machine Learning and Artificial Intelligence, Germany. This work has been funded by the Deutsche Forschungsgemeinschaft (DFG, German Research Foundation) under Germany's Excellence Strategy, EXC-2070 -- 390732324 (PhenoRob).
Corresponding: \texttt{jrueckin@uni-bonn.de}.}
\thanks{Digital Object Identifier (DOI): see top of this page.}
}
\begin{document}

\maketitle

\begin{abstract}
Semantic segmentation enables robots to perceive and reason about their environments beyond geometry. Most of such systems build upon deep learning approaches. As autonomous robots are commonly deployed in initially unknown environments, pre-training on static datasets cannot always capture the variety of domains and limits the robot's perception performance during missions. Recently, self-supervised and fully supervised active learning methods emerged to improve robotic vision. These approaches rely on large in-domain pre-training datasets or require substantial human labelling effort. We propose a planning method for semi-supervised active learning of semantic segmentation that substantially reduces human labelling requirements compared to fully supervised approaches. We leverage an adaptive map-based planner guided towards the frontiers of unexplored space with high model uncertainty collecting training data for human labelling. A key aspect of our approach is to combine the sparse high-quality human labels with pseudo labels automatically extracted from highly certain environment map areas. Experimental results show that our method reaches segmentation performance close to fully supervised approaches with drastically reduced human labelling effort while outperforming self-supervised approaches.
\end{abstract} 


\section{Introduction} \label{S:intro}

\IEEEPARstart{P}{erceiving} and understanding complex environments is a crucial prerequisite for autonomous systems~\citep{Lenczner2022, Georgakis2021}. At the same time, robots are increasingly utilised in diverse terrains to execute various tasks, such as monitoring~\citep{marchant2014sequential, hitz2017adaptive}, search and rescue~\citep{niroui2019deep, baxter2007multi}, and precision agriculture~\citep{popovic2020informative}. Thus, robotic perception systems need to adapt to novel domains and terrains. However, classical deep learning-based semantic segmentation systems are pre-trained on static datasets that often fall short in covering the varying domains and semantics encountered during real-world robot deployments.

This work examines the problem of semi-supervised active learning to improve robotic vision within an initially unknown environment. We aim to maximise the robot's semantic segmentation performance while minimising human labelling requirements. The robot re-plans paths online to collect informative training data to re-train a semantic segmentation model after a mission. We incorporate two sources of labels for network re-training based on the collected data: (i)~a human annotator and (ii)~automatically generated pseudo labels based on a semantic environment map incrementally built online during a mission.

\begin{figure}[!t]
    \captionsetup[subfigure]{labelformat=empty}
    \centering
    \subfloat[]{\includegraphics[width=\columnwidth]{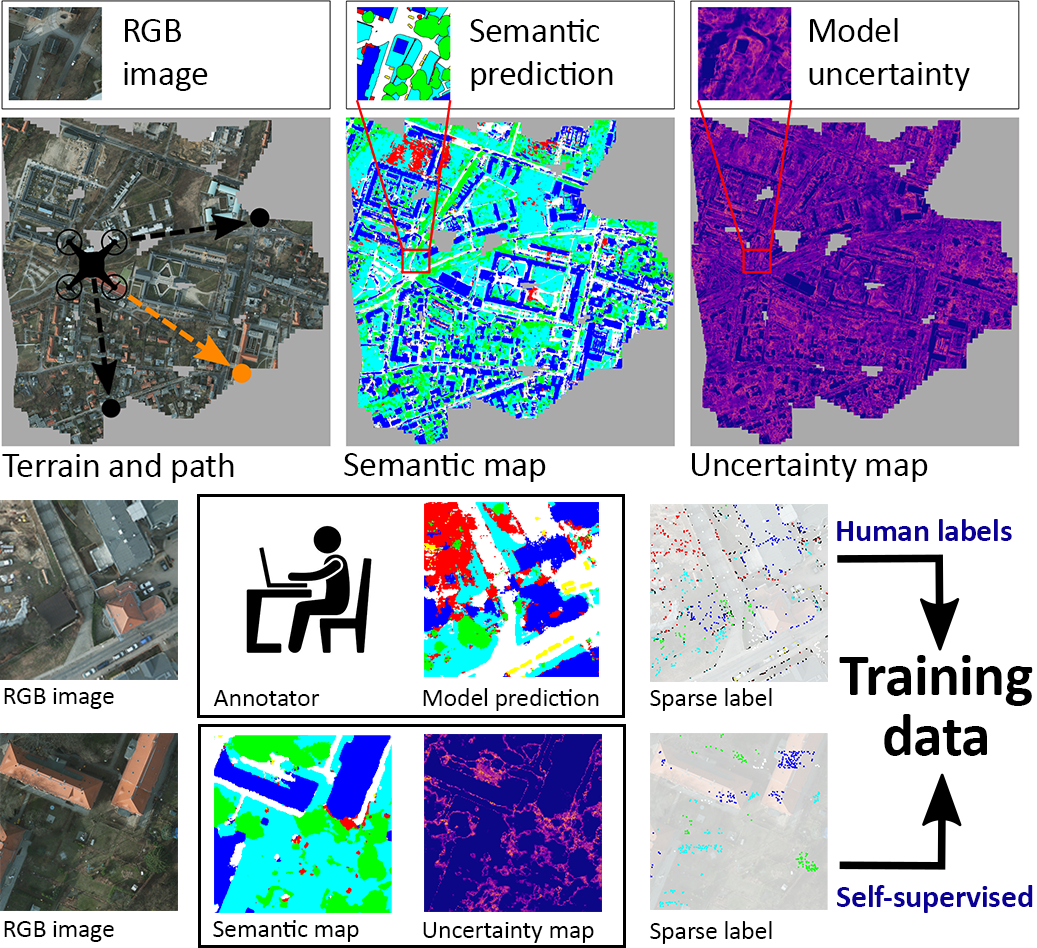}}
    \vspace{-3mm}
    \caption{Our semi-supervised active learning approach in an unknown environment (top). We infer semantic segmentation (top-centre) and model uncertainty (top-right) and fuse both in environment maps. The robot re-plans its path (orange, top-left) to collect diverse uncertain images. After each mission, we select sparse sets of pixels for human and self-supervised labelling (bottom). Self-supervised labels are rendered from low-uncertainty semantic map regions. Human labels are queried for regions of cluttered model predictions.}
    \label{F:teaser}
\end{figure}

To reduce human labelling effort, active learning methods select the most informative images from a static pool of unlabelled data~\citep{freund1997selective, gal2017deep, sener2017active, yang2017suggestive}. Recent works combine active learning with robotic planning to reduce the amount of labelled training data in unknown environments~\citep{Blum2019, ruckin2022informative, ruckin2023informative}. However, these methods require time-consuming dense pixel-wise human-labelled images to train semantic segmentation models. In parallel, self-supervised active learning methods automatically generate pseudo labels from semantic maps incrementally built during a mission~\citep{frey2021continual, zurbrugg2022embodied, chaplot2021seal}. Although these approaches do not rely on human labels, their applicability to unknown environments is limited since they require large labelled in-domain pre-training datasets to produce high-quality pseudo labels without systematic prediction errors.

The main contribution of this paper is a novel semi-supervised informative path planning approach for robotic active learning. Our approach bridges the gap between the general applicability of fully supervised methods and low human labelling requirements of self-supervised methods. A key novelty of our adaptive planning method is combining the selection of sparse and informative human-labelled training data and automatically generating highly certain pseudo labels as shown in \cref{F:teaser}. We fuse semantic predictions and Bayesian model uncertainty estimates~\citep{kendall2017uncertainties} into environment maps. Based on the model uncertainty map, our planner adaptively collects images from high-uncertainty areas. Following recent works in semi-supervised learning, we select only a sparse set of to-be-labelled informative pixels from each image~\citep{xie2022towards, shin2021all}. To further improve model performance, we automatically render highly certain pseudo labels based on the semantic and model uncertainty maps. By combining human and pseudo labels, we aim to maximise semantic segmentation performance while reducing human labelling effort compared to previous works in robotic active learning.

In sum, we make the following three key claims. First, our approach drastically reduces the number of human-labelled pixels compared to fully supervised active learning approaches while preserving similar semantic segmentation performance and outperforming self-supervised methods. Second, selecting sparse human labels in a targeted way improves semantic segmentation performance while minimising overall human labelling efforts. Third, the uncertainty-aware generation of pseudo labels further improves semantic segmentation performance compared to using human labels only. We will open-source our code for usage by the community at: \url{https://github.com/dmar-bonn/ipp-ssl}.

\section{Related Work} \label{S:related_work}

Our approach combines computer vision research aiming to minimise human labelling effort for training semantic segmentation models and informative path planning.

\textbf{Active learning} aims to select a minimal subset of informative to-be-labelled training data from a pool of unlabelled data that maximises model performance. Some works estimate uncertainty to access a sample's information value~\citep{gal2017deep, beluch2018power} utilising methods such as Monte-Carlo dropout~\citep{gal2017deep} or ensembles~\citep{beluch2018power}. These strategies select samples from a large pool of unlabelled data. In contrast, our planning method exploits Bayesian model uncertainty estimates~\citep{gal2017deep} fused into an environment map guiding the robot towards high-uncertainty areas to incrementally collect samples during a mission.

\citet{shin2021all} recently introduced an efficient label selection paradigm, which selects a sparse set of uncertain pixels for human labelling to train semantic segmentation models. \citet{benenson2022colouring} show that selecting sparse sets of to-be-labelled pixels reduces human labelling effort compared to dense pixel-wise human labels. Similarly, \citet{xie2022towards} propose a to-be-labelled pixel or region selection criterion for domain shift scenarios. In contrast to previous robotic planning methods~\citep{Blum2019, ruckin2022informative, ruckin2023informative}, which rely on dense pixel-wise human labels, our work utilises a new sparse human label selection strategy inspired by \citet{xie2022towards} to drastically reduce human labelling effort.

\textbf{Semi-supervised semantic segmentation} methods build upon a low budget of human-labelled training samples methods and improve model performance further by generating pseudo labels from model predictions of unlabelled data~\citep{he2021re, ouali2020semi}. Our work leverages a low number of sparsely human-labelled samples and combines them with automatically generated pseudo labels. In contrast to image-based pseudo label methods~\citep{ ouali2020semi, he2021re}, our approach renders pseudo labels from a semantic map in an uncertainty-aware fashion.

\textbf{Informative path planning} aims to maximise the collected information in initially unknown environments subject to robot constraints, such as mission time~\citep{ghaffari2018gaussian, chen2020autonomous, hitz2017adaptive}. Traditional  \textit{non-adaptive} approaches pre-compute static paths while \textit{adaptive} methods actively re-plan paths online based on collected measurements~\citep{hollinger2014sampling, hitz2017adaptive}. Our work focuses on \textit{adaptive} methods for active learning in semantic segmentation since they account for varying semantics and changing model uncertainties after network re-training.

Sampling-based techniques, such as receding-horizon planning for information gathering~\citep{hollinger2014sampling} or variants of Monte-Carlo tree search~\citep{ott2023sequential, marchant2014sequential, ruckin2022adaptive}, solve the informative path planning problem in a computationally efficient way. Similarly, optimisation-based strategies exploit algorithms such as the covariance matrix adaptation evolution strategy~\citep{hitz2017adaptive, popovic2020informative} to directly maximise objective functions. Geometric methods select potentially informative candidate robot poses at the frontiers of explored and unknown space~\citep{ghaffari2018gaussian, chen2020autonomous}. The above-mentioned works address informative path planning for classical exploration or monitoring tasks. In contrast, we develop a geometric planning approach to improve robot vision using semi-supervised active learning.

Recent works in active learning for semantic segmentation using robotic platforms follow either the paradigm of self-supervision without human labels~\citep{chaplot2021seal, zurbrugg2022embodied} or full human supervision for selected informative images requiring dense pixel-wise labels~\citep{Blum2019, ruckin2022informative, ruckin2023informative}. \citet{zurbrugg2022embodied} fuse 2D semantic predictions of a pre-trained network into a 3D map to automatically generate semantic labels for continual network re-training. Similarly, \citet{chaplot2021seal} train a viewpoint selection policy with reinforcement learning in simulation to target uncertain map parts. Despite not relying on human labels, these self-supervised methods require large human-labelled in-domain pre-training datasets in indoor scenes to produce high-quality pseudo labels. Further, systematic prediction errors prevent learning specific semantics~\citep{chaplot2021seal}. In contrast, \citet{Blum2019} propose a local planner to collect pixel-wise human-labelled data with high training data novelty~\citep{papernot2018deep}. \citet{ruckin2023informative} propose a general map-based planner for fully supervised active learning to improve semantic segmentation. While these works do not require large pre-training datasets, they still depend on dense pixel-wise human labels for model training.

Our approach combines the advantages of self- and fully supervised approaches into a semi-supervised adaptive informative path planning framework. We maintain the general applicability of fully supervised approaches while reducing human labelling efforts in active learning for robotic vision.

\section{Our Approach} \label{S:our_approach}

\begin{figure}[!t]
    \centering
    \includegraphics[width=\columnwidth]{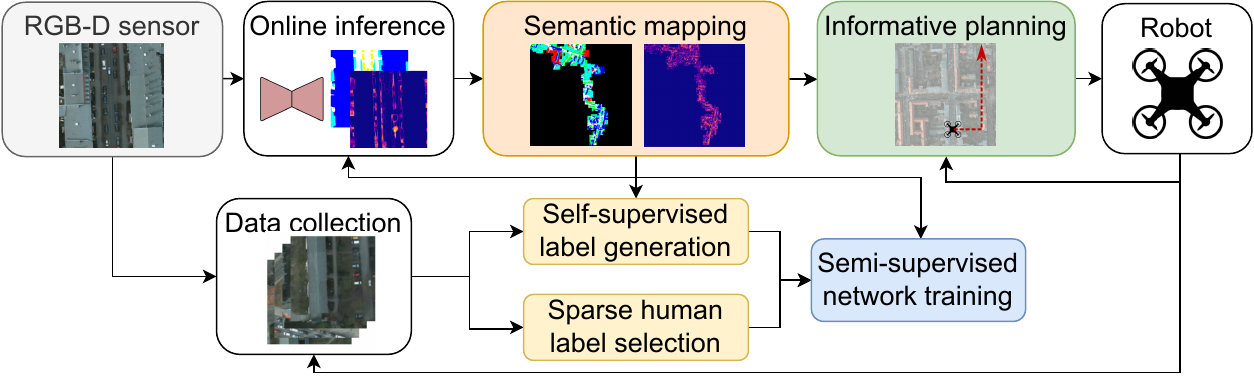}
    \caption{During a mission, a semantic segmentation network predicts pixel-wise semantics and model uncertainties from an RGB-D image. Both are fused into an uncertainty-aware semantic environment map (\Cref{SS:mapping}). Our planner guides the collection of training data for network re-training based on the robot state and map belief (\Cref{SS:planning}). After a mission, the collected data is labelled using two sources of labels: (i) a human annotator labels a sparse set of informative pixels, and (ii) we automatically render pseudo labels from the semantic map in an uncertainty-aware fashion.}
    \label{F:approach_overview}
\end{figure}

We present an adaptive informative path planning framework for semi-supervised active learning in semantic segmentation. Considering a robot equipped with an RGB-D sensor, our goal is to collect images in an initially unknown environment to improve semantic perception with minimal human labelling effort. \cref{F:approach_overview} summarises our framework. We predict pixel-wise semantics and associated model uncertainties to update a probabilistic semantic environment map (\cref{SS:mapping}). Based on the robot pose, flight budget, and map belief, we plan paths to adaptively collect training data in environment areas of high model uncertainty (\cref{SS:planning}). After a mission, we select a sparse set of informative to-be-human-labelled pixels in the collected images (\cref{SS:ssl_network_training}). We combine them with pseudo labels automatically rendered from the online-built semantic map in an uncertainty-aware fashion for network re-training (\cref{SS:self_supervised_label_generation}).

\subsection{Probabilistic Semantic Environment Mapping} \label{SS:mapping}

A crucial requirement for pseudo label generation and adaptive planning is a probabilistic map capturing information about the environment. We use a probabilistic multi-layered semantic environment mapping to fuse geometric and semantic information. The environment is discretised into three voxel maps $\mathcal{M}_{G}: V \to \{0,1\}^{W \times L \times H}, ~ \mathcal{M}_{S}: V \to \{0,1\}^{K \times W \times L \times H}, ~ \mathcal{M}_{U}: V \to [0,1]^{W \times L \times H}$ defined over $W \times L \times H$ spatially independent voxels $V$. $\mathcal{M}_{G}$ captures the geometric occupancy information, $\mathcal{M}_{S}$ stores the semantics, and $\mathcal{M}_{U}$ stores the associated model uncertainties. 

The semantic map $\mathcal{M}_{S}$ consists of $K$ layers with one layer per class. At each time step, a new RGB-D image arrives. The probabilistic semantic predictions and model uncertainties are inferred using a semantic segmentation model and Monte-Carlo dropout~\citep{ruckin2022informative}. We project the depth image, semantic predictions, and model uncertainties into the environment using the intrinsics and extrinsics of the RGB-D sensor. The geometric map $\mathcal{M}_{G}$ and semantic map $\mathcal{M}_{S}$ are recursively updated by probabilistic occupancy grid mapping~\citep{Elfes1989}. The model uncertainty map $\mathcal{M}_{U}$ is updated by maximum likelihood estimation.

Additionally, we maintain a count map $\mathcal{M}_{T}: V \to \mathbb{N}^{W \times L \times H}$ to track the occurrences in the human-labelled training data utilised in our planning objective. As the semantic segmentation model is re-trained after each robot mission, the semantic predictions and model uncertainties change. Following \citet{ruckin2023informative}, we re-compute the semantic and model uncertainty maps after model re-training using previously collected RGB-D images to obtain maximally up-to-date map priors for informative planning.

\subsection{Adaptive Informative Path Planning} \label{SS:planning}

Our planner is designed to collect new training data in initially unknown environments given mission budget constraints. We aim to maximise the performance of a semantic segmentation model with minimal human labelling effort after re-training it on the collected training data. Our planning method searches for a path $\psi^* = (\mathbf{p}_1,\ldots,\mathbf{p}_N) \in \Psi$ with a variable number $N \in \mathbb{N}$ of robot poses $\mathbf{p}_i \in \mathbb{R}^D$, $i \in \{1,\ldots,N\}$, in the set of potential paths $\Psi$, that maximises an information criterion $I: \Psi \to \mathbb{R}_{\geq 0}$:
\begin{equation} \label{eq:ipp_problem}
    \psi^* = \argmax_{\psi \in \Psi} I(\psi),\, \mathrm{s.t.}\, C(\psi) \leq B\,,
\end{equation}
where $I$ assigns an information value to each possible path $\psi \in \Psi$ and $B \geq 0$ is the mission budget. $C: \Psi \to \mathbb{R}_{\geq 0}$ defines the required budget to execute the path $\psi$.
%

At each time step $t$, we adaptively re-plan the next-best robot pose $\mathbf{p}^{*}_{t+1}$ to collect informative training data. We utilise a geometric frontier-based planner~\citep{yamauchi1997frontier, ruckin2023informative} guided by the information criterion $I$. The information criterion estimates the effect of a candidate training image recorded at a robot pose on a semantic segmentation model's performance. Based on the geometric map belief $\mathcal{M}_{G}^t$, we assign each voxel $v \in V$ to one of three disjoint sets of voxels $V_F \cup V_U \cup V_O = V$ containing the free, unknown, and surface voxels, respectively.
%
To generate potentially informative robot pose candidates $\mathbf{p}^{c}_{t+1} \in \mathbb{R}^D$, we equidistantly sample poses $\mathbf{p}^{c}_{t+1}$ along the frontiers of free and unknown space reachable within the remaining mission budget. A frontier is a set of connected free voxels in $V_F$ with neighbouring unknown voxels in $V_U$~\citep{yamauchi1997frontier}. The information value $I(\mathbf{p}^{c}_{t+1})$ of a candidate pose $\mathbf{p}^{c}_{t+1}$ is defined as~\citep{zurbrugg2022embodied}:
\begin{equation} \label{eq:information_criterion}
    I(\mathbf{p}^{c}_{t+1}) = \sum_{v \in \mathrm{Img}(\mathbf{p}^{c}_{t+1})}
    \begin{cases}
        0 & \text{, if} ~ v \in V_F \\
        c_u & \text{, if} ~ v \in V_U \\
        \frac{\mathcal{M}_{U}^t(v)}{\mathcal{M}_{T}^t(v)} & \text{, if} ~ v \in V_O\, ,
    \end{cases}
\end{equation}
where $c_u \in \mathbb{R} \geq 0$ is a uniform model uncertainty prior fostering exploration of unobserved environment areas, and $\mathrm{Img}(\mathbf{p}^{c}_{t+1})$ is a rendered 2D image of voxels visible from $\mathbf{p}^{c}_{t+1}$ with resolution $w' \times h'$. We obtain $\mathrm{Img}(\mathbf{p}^{c}_{t+1})$ by ray casting into the geometric map belief $\mathcal{M}_G^t$ from pose $\mathbf{p}^{c}_{t+1}$. 
While casting a ray from $\mathbf{p}^{c}_{t+1}$, only free voxels are treated as traversable. Unknown or occupied voxels are assumed to be reflective. Pixels corresponding to rays that traverse only free voxels are assigned zero information value as we assume semantics are only assigned to surfaces. If a surface voxel reflects a ray, its effect on semantic segmentation performance after model re-training is estimated by its model uncertainty. To trade-off between model uncertainty and training data diversity, we normalise a voxel's information value by its number of occurrences in the training dataset. 

\subsection{Semi-Supervised Training} \label{SS:ssl_network_training}

The main contribution of our approach is a semi-supervised training strategy for improving the robot's semantic perception. We utilise a semantic segmentation network $\mathbf{f_{\theta}}(\mathbf{z}) = p(\mathbf{y}\,|\,\mathbf{z}) \in [K \times w \times h]$ parameterised by $\theta$ to predict the pixel-wise probabilities of $K$-class semantic labels $\mathbf{y} \in \{1,\ldots,K\}^{w \times h}$ given an input RGB image $\mathbf{z}$ of resolution $w \times h$. We follow \citet{ruckin2022informative} to estimate pixel-wise model uncertainties $\mathbf{u} \in [0,1]^{w \times h}$ via Monte-Carlo dropout~\citep{kendall2017uncertainties}. To maximise model performance, we combine human labels $\mathbf{Y}_l = \{\mathbf{y}_l^1,\ldots,\mathbf{y}_l^{N_l}\}$ of images $\mathbf{Z}_l = \{\mathbf{z}_l^1,\ldots,\mathbf{z}_l^{N_l}\}$ with pseudo labels $\mathbf{Y}_u = \{\mathbf{y}_u^1,\ldots,\mathbf{y}_u^{N_u}\}$ of images $\mathbf{Z}_u = \{\mathbf{z}_u^1,\ldots,\mathbf{z}_u^{N_u}\}$, where $N_l$ and $N_u$ are the numbers of human-labelled and pseudo-labelled images. To reduce human labelling effort, we consider a sparse set of human-labelled pixels $\mathbf{y}_l^i$ and pseudo-labelled pixels $\mathbf{y}_u^i$. Each non-labelled pixel in some label $\mathbf{y}_l^i$ or $\mathbf{y}_u^i$ is assigned a void class $N^v \in \{1,\ldots,K\}$. During training, we mask the loss with $\mathbb{I}_{\mathbf{y} \neq N^v} \in \{0,1\}^{w \times h}$, where $\mathbb{I}_{\mathbf{y} \neq N^v}$ is zero for each pixel with class $N^v$. The model~$f_{\theta}$ is trained to minimise the following cross-entropy loss function:
\begin{equation} \label{eq:network_loss}
    \begin{aligned}
        \mathcal{L}(\theta) = &\frac{1}{N_l \alpha} \sum_{i=1}^{N_l} -\log \left(\mathbf{f_{\theta}}(\mathbf{z}_l^i)^{(\mathbf{y}_l^i, :, :)} \right) \mathbb{I}_{\mathbf{y}_l^i \neq N^v} + \\ &\frac{1}{N_u \alpha} \sum_{i=1}^{N_u} -\log \left(\mathbf{f_{\theta}}(\mathbf{z}_u^i)^{(\mathbf{y}_u^i, :, :)} \right) \mathbb{I}_{\mathbf{y}_u^i \neq N^v} 
    \end{aligned}
\end{equation}
where $\alpha \in \mathbb{N}$ is the number of labelled pixels per image
and $\mathbf{f_{\theta}}(\mathbf{z})^{(\mathbf{y}, :, :)}$ are the probabilities of ground truth semantics $\mathbf{y}$.

Combining ideas from \citet{shin2021all} and \citet{xie2022towards}, we propose a new model architecture-agnostic pixel selection procedure for sparse human labels that trades off between label informativeness and diversity. After each mission, for all newly collected images $\mathbf{z}_l$ recorded at planned poses maximising \Cref{eq:information_criterion}, for each pixel $(m,n)$, we predict semantic probabilities $p(\mathbf{y}\,|\,\mathbf{z}_l)^{(:, m,n)}$ and extract the maximum likelihood label $\Tilde{\mathbf{y}_l}^{(m,n)} = \argmax_{k \in [K]} p(\mathbf{y}\,|\,\mathbf{z}_l)^{(k,m,n)}$. We compute each pixel's region impurity score~\citep{xie2022towards} as follows:
\begin{equation} \label{eq:region_impurity}
    \begin{aligned}
        R_r(\mathbf{z}_l)^{(m,n)} = -\sum_{k=1}^{K} \log \left(\frac{\lvert N_r^k(m,n) \rvert}{(2r+1)^2}\right) \frac{\lvert N_r^k(m,n) \rvert}{(2r+1)^2},\\
        N_r^k(m,n) = \left\{(i,j) \in N_r(m,n)\,|\,\Tilde{\mathbf{y}_l}^{(i,j)}=k\right\},
    \end{aligned}
\end{equation}
where $N_r(m,n) = \{(i,j)\,|\,\lvert i-m \rvert \leq r, \lvert j-n \rvert \leq r \}$ is the set of $r$-step neighboring pixels of $(m,n)$. Intuitively, the region impurity for human labelling a pixel is high whenever the number of different classes predicted within the pixel's $r$-step neighbourhood is high, as a well-trained model should predict locally non-cluttered semantics. In contrast to \citet{xie2022towards}, we do not greedily select the $\alpha$ pixels per image that maximise region impurity. Instead, per image, we sample $\alpha$ pixels uniformly at random from the $\beta\,\%$ pixels with the highest region impurity to foster human label diversity. While $\alpha$ sets the user-defined human labelling budget, $\beta$ implicitly provides a lower bound for a pixel's information value. Experimentally, we found that smaller values $\beta \leq 10\,\%$ ensure informative pixel selection, while $\beta \to 100\,\%$ lead to inefficient random pixel selection. Further, both region impurity and random sampling are crucial for maximising model performance.

\subsection{Self-Supervised Pseudo Label Generation} \label{SS:self_supervised_label_generation}

Similarly to self-supervised robotic active learning approaches~\cite{frey2021continual, zurbrugg2022embodied}, after a mission is finished, we utilise our incrementally online-built uncertainty-aware semantic map (\Cref{SS:mapping}) to generate pseudo labels $\mathbf{Y}_u$ in a self-supervised fashion. We record to-be-pseudo-labelled images $\mathbf{z}_u \in \mathbf{Z}_u$ equidistantly between two poses planned for collecting to-be-human-labelled images (\Cref{eq:information_criterion}) to maximise training data diversity. Given a robot pose $\mathbf{p}_u \in \mathbb{R}^D$ at which $\mathbf{z}_u$ is recorded, we render a pixel-wise probabilistic pseudo label $p(\mathbf{y}_u\,|\,\mathbf{p}_u, \mathcal{M}_S) \in [0,1]^{K \times w \times h}$ from the semantic map belief $\mathcal{M}_S$ at the image resolution $w \times h$. Then, for each pixel $(m,n)$, we extract the maximum likelihood pseudo label $\mathbf{y}_u^{(m,n)} = \argmax_{k \in [K]} p(\mathbf{y}_u\,|\,\mathbf{p}_u, \mathcal{M}_S)^{(k,m,n)}$. Similarly, we render the corresponding pixel-wise model uncertainty $\mathbf{u}_u \in [0,1]^{w \times h}$ from the model uncertainty map $\mathcal{M}_U$. If a ray corresponding to pixel $(m,n)$ is not reflected by a surface voxel in $\mathcal{M}_G$, we assign the void class $\mathbf{y}_u^{(m,n)} = N^v$.

In contrast to previous works~\citep{frey2021continual, zurbrugg2022embodied}, we only use a sparse set of $\alpha$ pseudo-labelled pixels per image $\mathbf{z}_u$ to train the network via \Cref{eq:network_loss} to balance the human and pseudo label supervision. We extend the approach of \citet{shin2021all} to a new pixel selection procedure for sparse pseudo labels $\mathbf{y}_u$ that trades off between semantic map uncertainty and pseudo label diversity. After each mission, for all images~$\mathbf{z}_u$ collected in any of the previous missions, we (re-)render pseudo labels $\mathbf{y}_u$ and model uncertainties $\mathbf{u}_u$ based on the most recent map beliefs $\mathcal{M}_S$ and $\mathcal{M}_U$. Similar to the human-labelled pixel selection (\Cref{SS:ssl_network_training}), for each image, we sample $\alpha$ pixels $(m,n)$ at random from the $\beta \%$ pixels with the lowest map-based model uncertainty $\mathbf{u}_u^{(m,n)}$. Non-sampled pixels are assigned the void class. We found that providing an implicit upper bound $\beta$ for model uncertainty yields higher semantic segmentation performance than random sampling as it acts as a proxy to the pseudo label quality. Further, $\beta \leq 10\,\%$ usually ensures moderate model performance improvements.

\section{Experimental Results} \label{S:experimental_results}

Our experiments are designed to assess the performance of our approach. They support the claims made in this paper. First, we show that our method for selecting human-labelled pixels outperforms state-of-the-art pixel selection methods in our robotic planning context (\Cref{SS:exp_human_label_selection}). Second, we validate that combining our uncertainty-aware pseudo labels with human labels improves semantic segmentation performance and drastically reduces the number of human-labelled pixels compared to fully supervised approaches while maintaining similar performance (\Cref{SS:exp_pseudo_label_generation}). Third, our semi-supervised active learning approach outperforms self-supervised active learning approaches (\Cref{SS:exp_ssl_for_al}).

\subsection{Experimental Setup} \label{SS:exp_setup}

\textbf{Baseline \& Dataset.} We compare our frontier planner against a coverage-based strategy that pre-computes paths to maximise spatial coverage~\citep{ruckin2023informative}.
We evaluate our approach on the real-world $7$-class ISPRS Potsdam orthomosaic dataset~\cite{Potsdam2018} and simulate $10$ subsequent \ac{UAV} missions from $30$\,m altitude with a mission budget of $1800$\,s. The \ac{UAV} uses a downwards-facing RGB-D camera with a footprint of $400$\,px$\times 400$\,px~\citep{ruckin2023informative}.

\textbf{Evaluation Metrics.} We evaluate semantic segmentation performance (dependent variable) over the number of human-labelled training images or pixels (independent variable). We use \ac{mIoU}~\citep{cordts2016cityscapes} and pixel-wise accuracy~\citep{deng2009imagenet} to quantify semantic segmentation performance.
We run three trials per experiment and report the mean and standard deviation performance curves.

\textbf{Implementation Details.} We use Bayesian ERFNet~\citep{ruckin2022informative} pre-trained on the Cityscapes dataset~\citep{cordts2016cityscapes}. Re-training after each mission starts from this checkpoint. The model is trained until convergence on the validation set. We use a one-cycle learning rate scheduler, a batch size of $8$, and weight decay $\lambda = (1 - p) / 2N$, where $p = 0.5$ is the dropout probability, and $N = N_l + N_u$ is the number of training images~\cite{gal2017deep}.
The human and pseudo label pixel selection lower bounds are $\beta=5\%$, and the $r$-neighborhood of the human label selection criterion is set to $r=1$. In practice, our approach allows for using any user-defined model.

\subsection{Targeted Human Label Selection} \label{SS:exp_human_label_selection}

\begin{figure}[!t]
    \centering
    \includegraphics[width=\columnwidth]{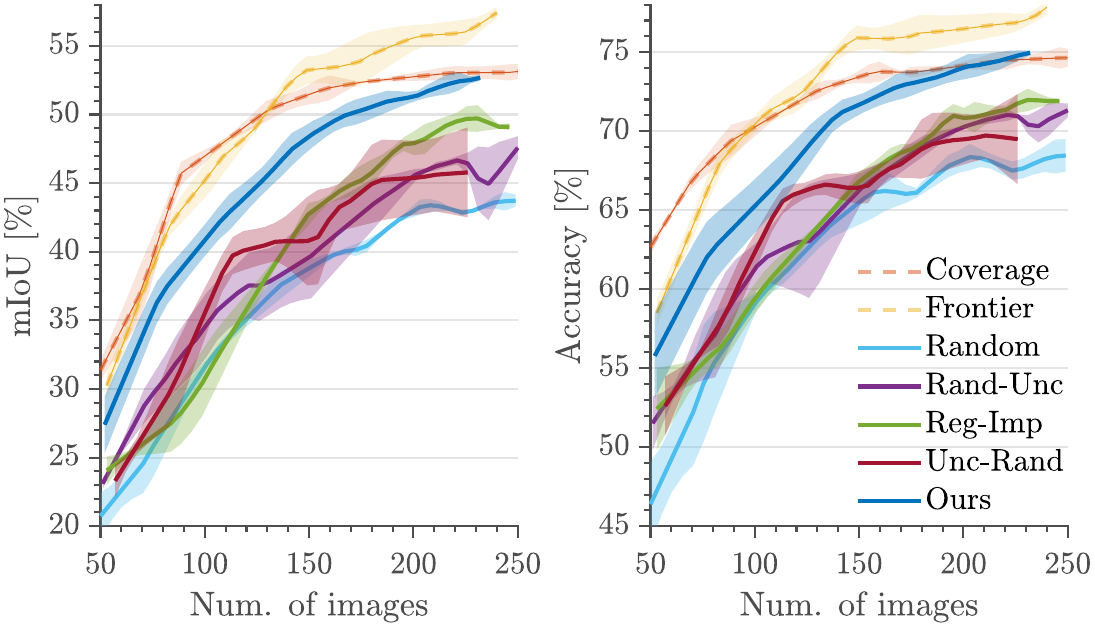}
    \caption{Comparison of label selection methods with $\alpha = 1000$\, human-labelled pixels per image using our frontier planner on ISPRS Potsdam. Frontier (yellow) and coverage (orange) planners use densely labelled images indicating performance upper bounds. Results are averaged over three runs. Shaded regions indicate one standard deviation. Our proposed method (dark blue) outperforms state-of-the-art pixel selection methods.}
    \label{F:sparse_human_labels_1000_pixels}
\end{figure}

\begin{table}[!t]
\centering
\caption{Per-class IoU comparison of sparse label selection methods with $\alpha = 1000$\, human-labelled pixels per image using our frontier planner on ISPRS Potsdam. \textit{Dense} uses dense pixel-wise human-labelled images indicating the performance upper bound.}
\label{T:sparse_human_labels_per_class_1000_pixels}
\begin{tabular}{@{}lccccccc@{}}
\toprule
Method       & \rotatebox[origin=c]{70}{Mission}            & \rotatebox[origin=c]{70}{Surface}        & \rotatebox[origin=c]{70}{Building}       & \rotatebox[origin=c]{70}{Vegetation}     & \rotatebox[origin=c]{70}{Tree}           & \rotatebox[origin=c]{70}{Car}            & \rotatebox[origin=c]{70}{Clutter}     \\ \toprule
Random       & \multirow{4}{*}{3} & 53.98          & 51.00          & 40.58          & 20.87          & 28.54          & 7.58           \\
Unc-Rand     &                    & 58.93          & 60.89          & 43.09          & 25.15          & 42.04          & 11.42          \\
Reg-Imp      &                    & 51.17          & 48.84          & 39.96          & 15.29          & 0.00           & 8.26            \\
\textbf{Ours} &                    & \textbf{59.47} & \textbf{65.74} & \textbf{46.37} & \textbf{33.74} & \textbf{47.20} & \textbf{15.36} \\ \midrule
Dense        &                    & 63.93          & 70.39          & 49.46          & 35.82          & 60.95          & 10.40         \\ \toprule
Random       & \multirow{4}{*}{6} & 59.16          & 63.30          & 43.33          & 31.62          & 44.68          & 11.63          \\
Unc-Rand     &                    & 61.87          & 68.99          & 42.50          & 29.80          & 52.57          & \textbf{16.60} \\
Reg-Imp      &                    & 60.19          & 69.61          & 46.68          & 30.83          & 59.49          & 12.59          \\
\textbf{Ours} &                    & \textbf{65.99} & \textbf{72.83} & \textbf{51.56} & \textbf{41.16} & \textbf{61.07} & 15.53          \\ \midrule
Dense        &                    & 71.08          & 77.72          & 53.14          & 45.80          & 68.81          & 17.56          \\ \toprule
Random       & \multirow{4}{*}{9} & 59.38          & 64.71          & 43.80          & 33.21          & 50.54          & 11.33          \\
Unc-Rand     &                    & 62.40          & 70.19          & 46.68          & 30.91          & 57.32          & 14.92          \\
Reg-Imp      &                    & 62.31          & 71.78          & 46.87          & 36.92          & 64.57          & 12.33          \\
\textbf{Ours} &                    & \textbf{67.94} & \textbf{74.54} & \textbf{52.00} & \textbf{43.37} & \textbf{66.50} & \textbf{16.67} \\ \midrule
Dense        &                    & 71.23          & 78.60          & 52.79          & 48.52          & 71.57          & 20.11          \\ \bottomrule
\end{tabular}
\end{table}

\begin{figure}[!t]
    \captionsetup[subfigure]{labelformat=empty}

    \centering
    \subfloat[]{\includegraphics[width=0.19\columnwidth]{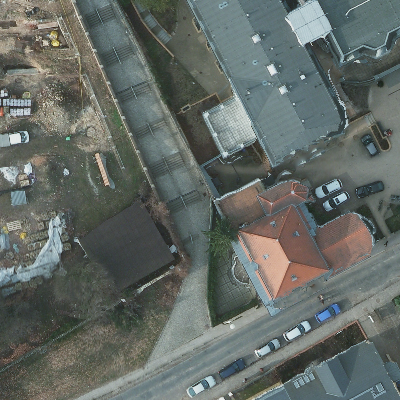}}
    \hfill
    \subfloat[]{\includegraphics[width=0.19\columnwidth]{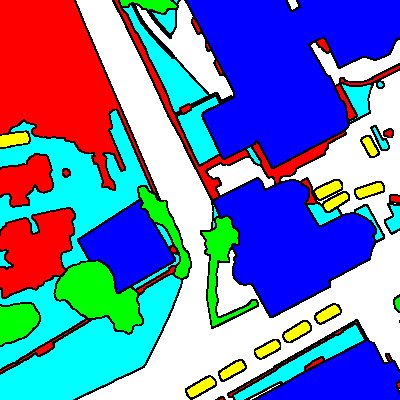}}
    \hfill
    \subfloat[]{\includegraphics[width=0.19\columnwidth]{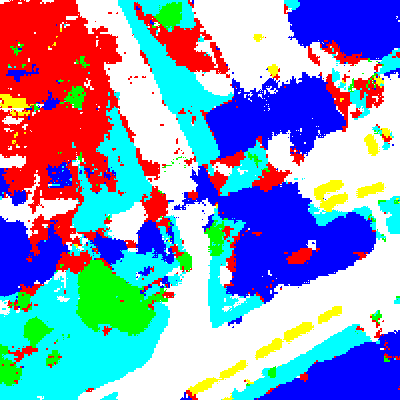}}
    \hfill
    \subfloat[]{\includegraphics[width=0.19\columnwidth]{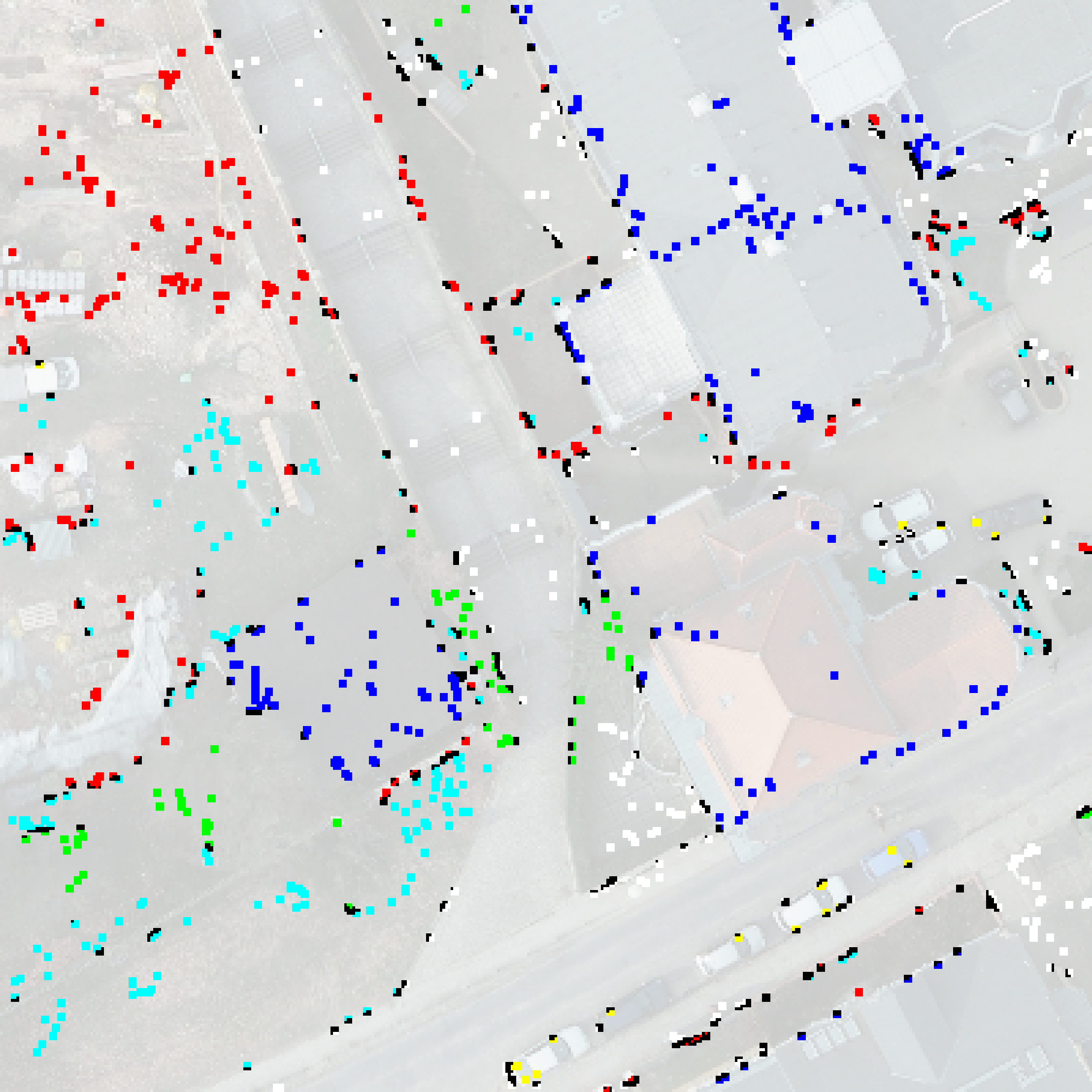}}
    \hfill
    \subfloat[]{\includegraphics[width=0.19\columnwidth]{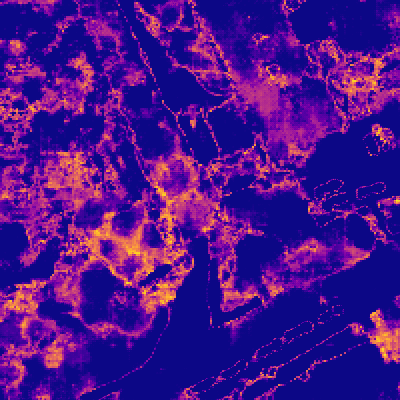}}

    \vspace{-4mm}
    \subfloat[]{\includegraphics[width=0.19\columnwidth]{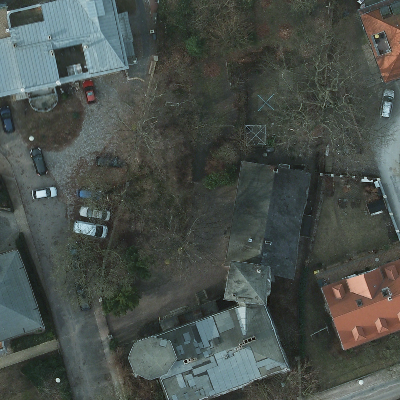}}
    \hfill
    \subfloat[]{\includegraphics[width=0.19\columnwidth]{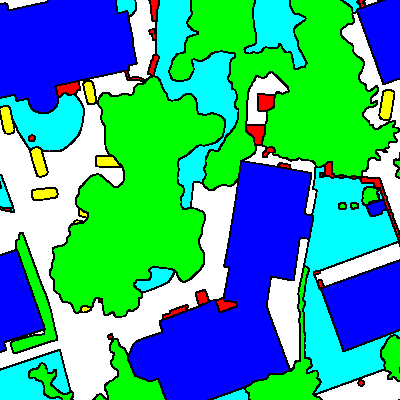}}
    \hfill
    \subfloat[]{\includegraphics[width=0.19\columnwidth]{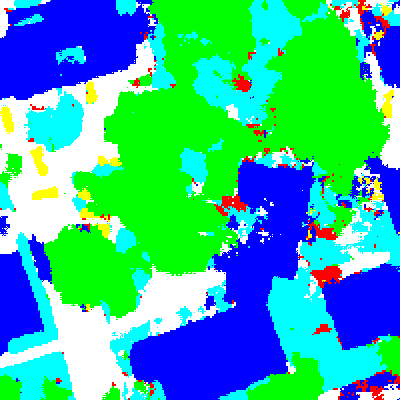}}
    \hfill
    \subfloat[]{\includegraphics[width=0.19\columnwidth]{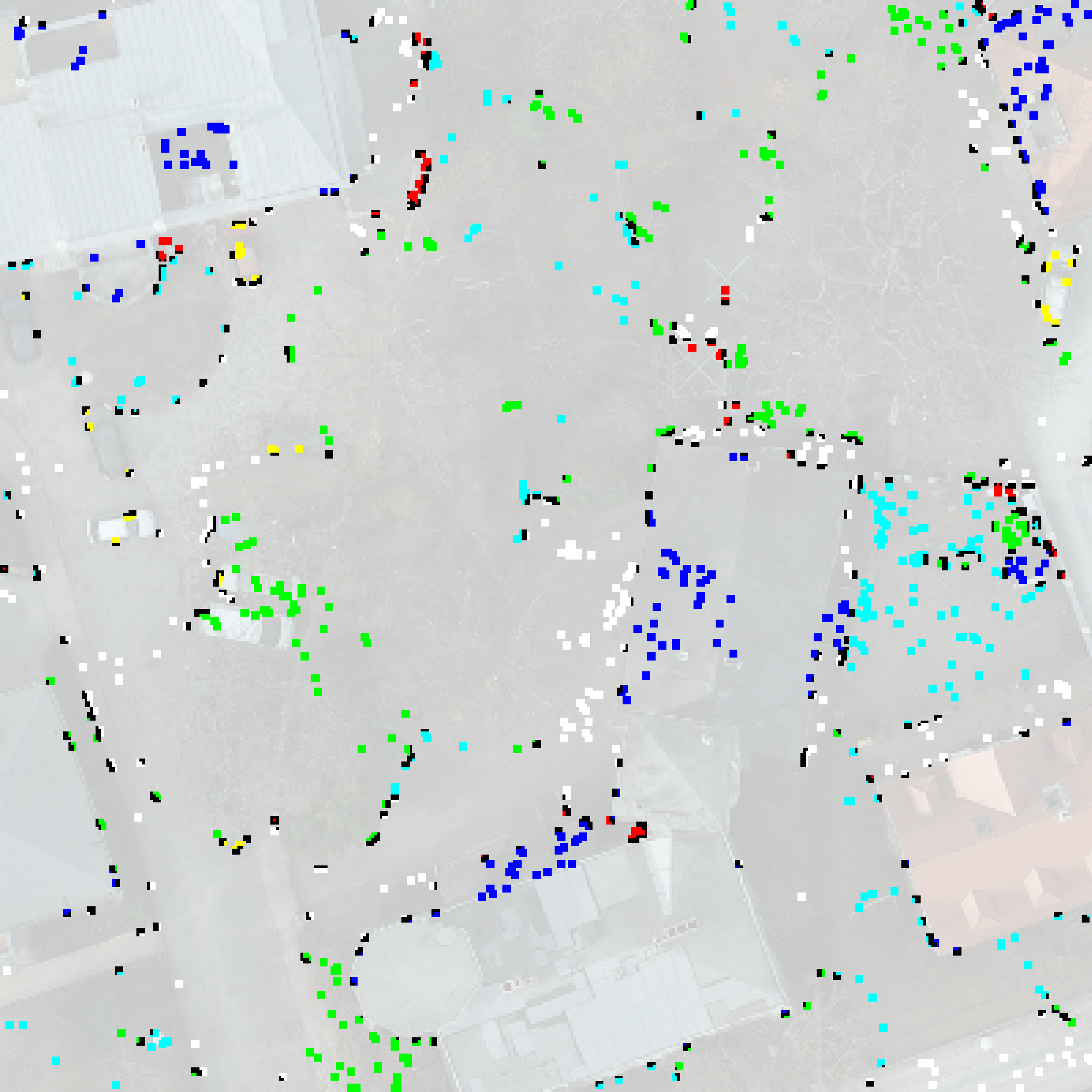}}
    \hfill
    \subfloat[]{\includegraphics[width=0.19\columnwidth]{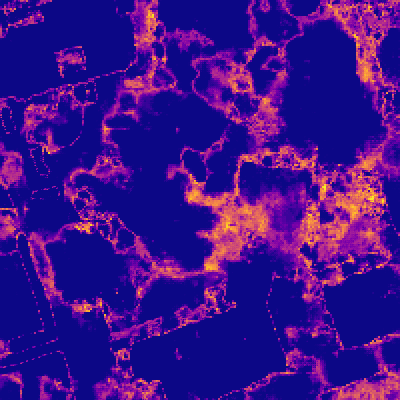}}

    \vspace{-4mm}
    \subfloat[Input]{\includegraphics[width=0.19\columnwidth]{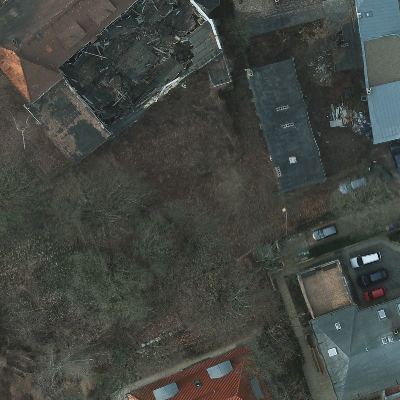}}
    \hfill
    \subfloat[Ground truth]{\includegraphics[width=0.19\columnwidth]{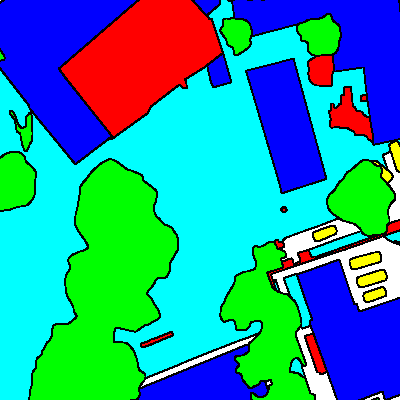}}
    \hfill
    \subfloat[Prediction]{\includegraphics[width=0.19\columnwidth]{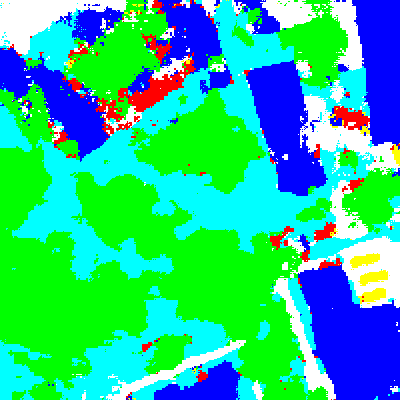}}
    \hfill
    \subfloat[Sparse Label]{\includegraphics[width=0.19\columnwidth]{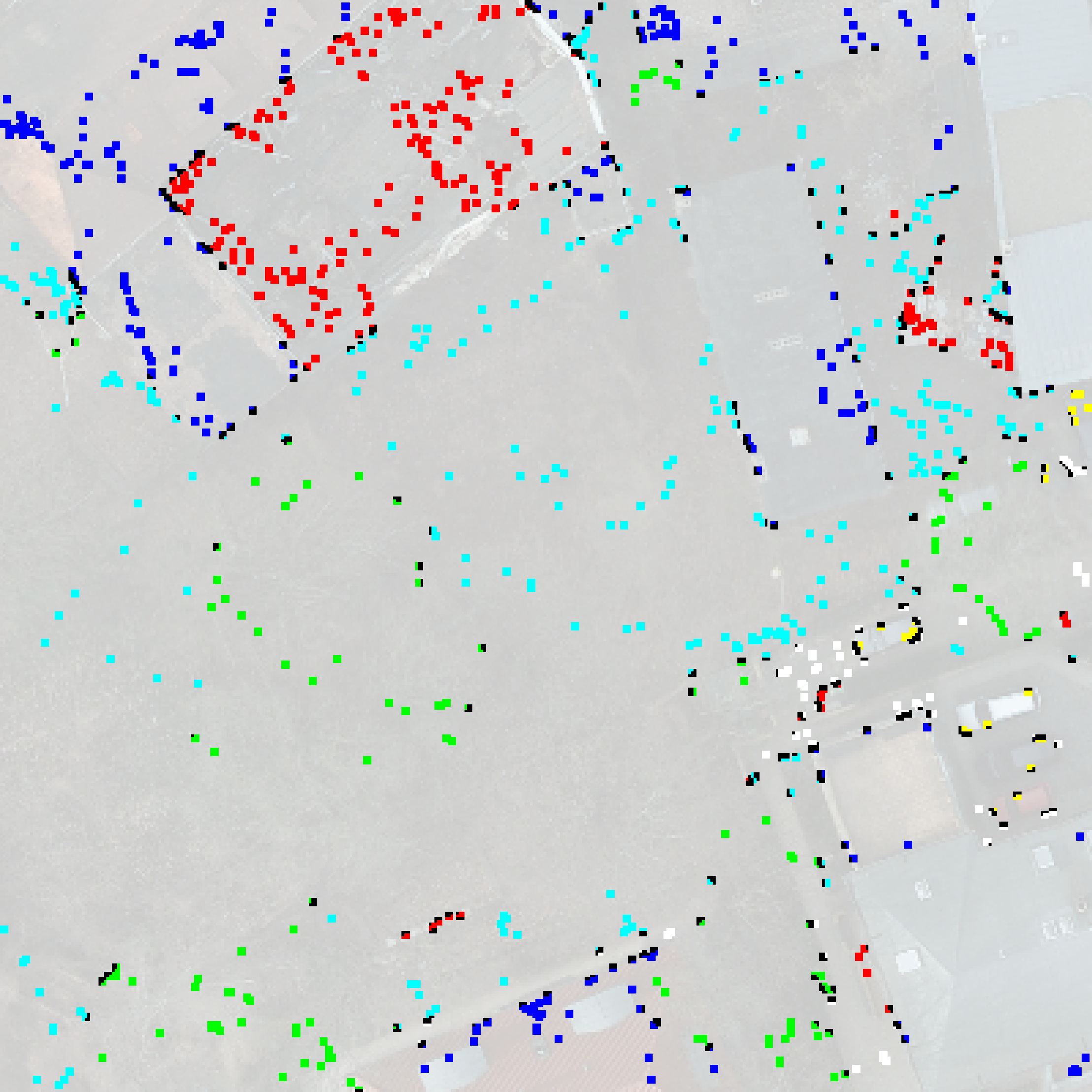}}
    \hfill
    \subfloat[Uncertainty]{\includegraphics[width=0.19\columnwidth]{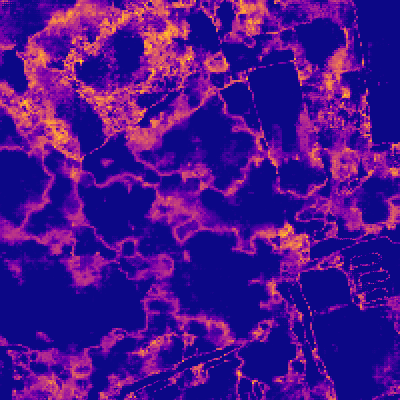}}

\caption{Qualitative results of our human label pixel selection method on ISPRS Potsdam. Columns from left to right: RGB input, ground truth, prediction, pixels selected for re-training, model uncertainty. Selected pixels are expanded to their one-pixel neighbourhood for visualisation. Our method selects pixels in areas of cluttered predictions, often corresponding to misclassified regions.}
\label{F:sparse_human_labels_qualitative_results}
\end{figure}

The first set of experiments shows that targeted human label selection improves semantic segmentation performance and reduces human labelling effort. We verify that our method (i) outperforms state-of-the-art pixel selection methods in the robotic planning context and (ii) improves semantic segmentation performance over non-targeted pixel selection with higher gains for lower human labelling budgets. The experiments are conducted using human labels only. 

We compare our human-labelled pixel selection method (\textit{Ours}, \Cref{SS:ssl_network_training}) against four pixel selection methods for a low human labelling budget of $\alpha = 1000 \approx 0.6\%$ pixels for each collected image by our frontier planner (\Cref{SS:planning}). Namely: (i) sample $\alpha$ pixels from the $\beta \%$ most uncertain pixels~\citep{shin2021all} (\textit{Unc-Rand}); (ii) sample $\beta \%$ pixels at random, then select the $\alpha$ most uncertain pixels~\citep{shin2021all} (\textit{Rand-Unc}); (iii) select $\alpha$ pixels uniformly at random (\textit{Random}); and (iv) select $\alpha$ pixels with the highest region impurity in an $r$-neighborhood~\citep{xie2022towards} (\textit{Reg-Imp}), where $r=1$ yields the best results. Additionally, we show results for the \textit{Frontier} and \textit{Coverage} planner utilising pixel-wise densely human-labelled images~\citep{ruckin2023informative} as an upper performance bound. 

\begin{figure}[!t]
    \centering
    \includegraphics[width=\columnwidth]{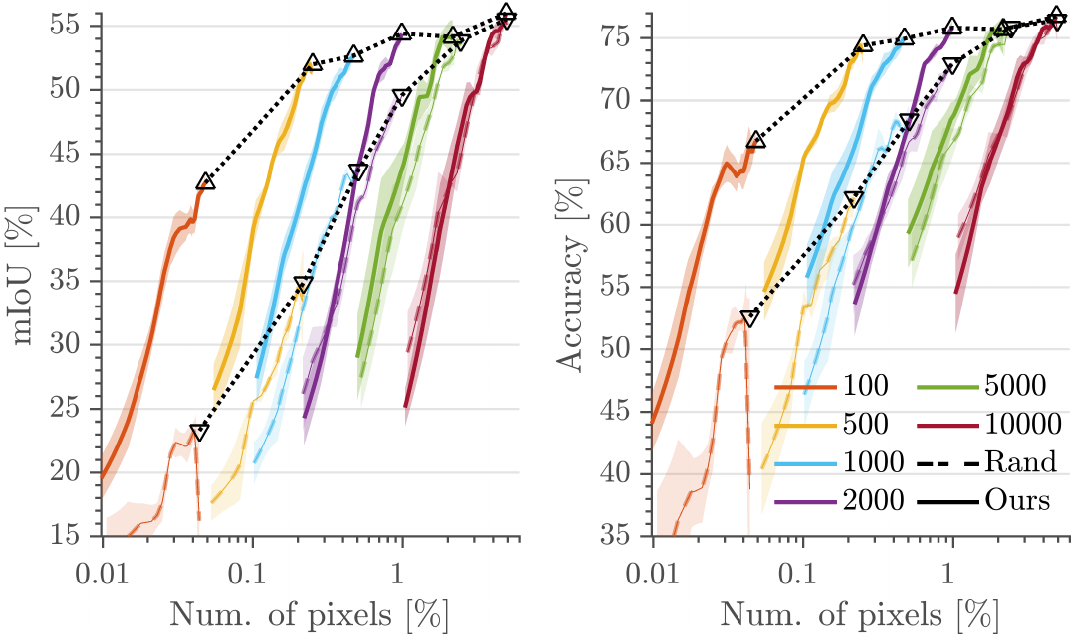}
    \caption{Comparison of our human label selection method (solid lines) to random label selection (dashed transparent lines) over varying labelling budgets $\alpha \in [100, 10000]\,\text{px}$ using our frontier planner on ISPRS Potsdam. Results are averaged over three runs. Shaded regions indicate one standard deviation. The performance gain of our method drastically increases for lower labelling budgets.}
    \label{F:sparse_human_labels_varying_budget}
\end{figure}

\cref{F:sparse_human_labels_1000_pixels} summarises the semantic segmentation performance of the different pixel selection methods. In line with previous fully supervised approaches~\citep{Blum2019, ruckin2023informative}, the \textit{Frontier} planner (yellow) using densely labelled images achieves the highest performance outperforming the non-adaptive \textit{Coverage} planner (orange). Notably, our method (dark blue) shows the fastest improvement and highest final \ac{mIoU} of $\approx 52.5\%$ of all pixel selection methods, significantly outperforming the second-best \textit{Reg-Imp} method (green) reaching $\approx 49\%$ final \ac{mIoU}. Particularly, our human label selection matches the final performance of the \textit{Coverage} planner using only $\approx 0.6\%$ of the labelled pixels. \cref{T:sparse_human_labels_per_class_1000_pixels} shows the superior per-class performance of our human label selection method, verifying its ability to select sparse but informative human labels for different semantics. \cref{F:sparse_human_labels_qualitative_results} displays images collected during a mission, onboard semantic predictions, and corresponding human-labelled pixels selected with our method.

\cref{F:sparse_human_labels_varying_budget} shows the semantic segmentation performance of our targeted pixel selection method (solid lines) compared to randomly selecting human-labelled pixels (dashed transparent lines) over varying human labelling budgets. Noticeably, for budgets $\alpha \leq 2000\,\text{px} \approx 1.3\,\%$, our pixel selection method clearly outperforms random pixel selection. Favourably, the performance gain of our pixel selection method over random pixel selection drastically increases with lower human labelling budgets. For an extremely low budget of $\alpha = 100\,\text{px} \approx 0.06\,\%$, our targeted pixel selection method leads to a high final performance gain of $\approx 20\,\%$ \ac{mIoU}.

\subsection{Uncertainty-Aware Pseudo Label Generation} \label{SS:exp_pseudo_label_generation}

\begin{figure}[!t]
    \centering
    \includegraphics[width=\columnwidth]{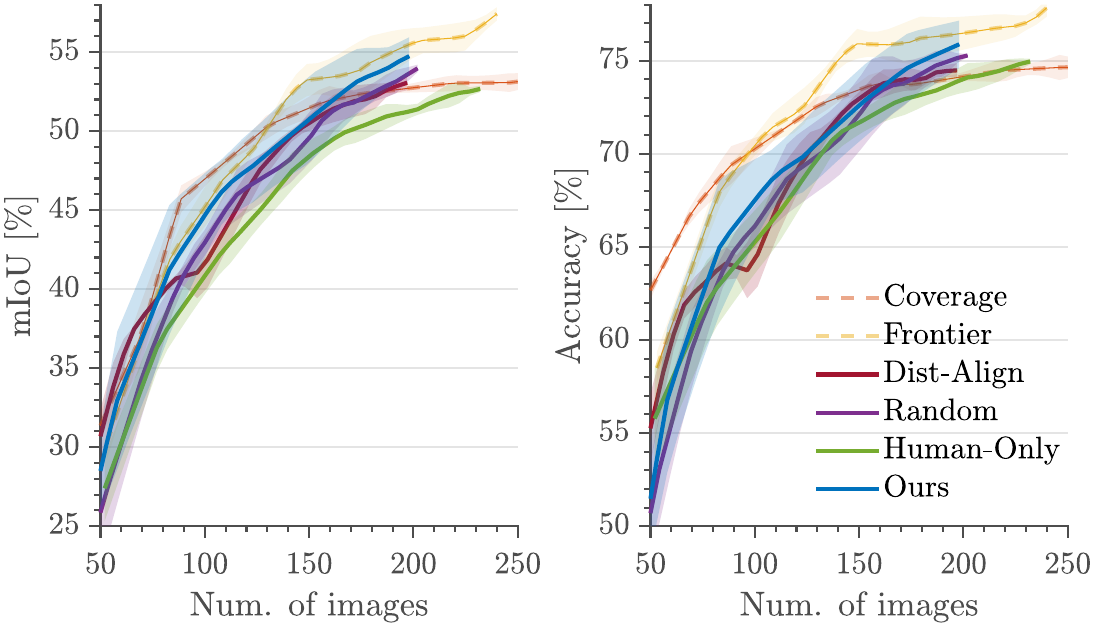}
    \caption{Comparison of pseudo label selection methods with $\alpha = 1000$\, human- and pseudo-labelled pixels per image using our frontier planner on ISPRS Potsdam. Frontier (yellow) and coverage (orange) planners use densely labelled images indicating performance upper bounds. Results are averaged over three runs. Shaded regions indicate one standard deviation. Our method (dark blue) outperforms other pseudo label selection methods.}
    \label{F:pseudo_human_labels_combined_1000_pixels}
\end{figure}

The second set of experiments shows that uncertainty-aware generation of pseudo labels improves semantic segmentation performance. We validate that (i) our pseudo label selection method outperforms other selection strategies, (ii) combining our human label selection with our pseudo label selection consistently improves semantic segmentation performance across varying labelling budgets, and (iii) our semi-supervised approach drastically reduces the number of human-labelled pixels compared to fully supervised approaches while maintaining similar performance. The experiments are conducted using our human label selection method.

We compare our uncertainty-aware pseudo label selection (\textit{Ours}, \Cref{SS:self_supervised_label_generation}) against two other pseudo label selection methods for a low human labelling budget of $\alpha = 1000\,\text{px} \approx 0.6\,\%$ per image. Namely: (i) we re-distribute the pseudo labels' class distribution to the true class distribution estimated by the human labels using per-class model uncertainty thresholds to select on average $\alpha$ pixels per image~\cite{he2021re} (\textit{Dist-Align}), and (ii) we randomly select $\alpha$ pixels per image (\textit{Random}). We compare against using $\alpha$ human-labelled pixels per image only (\textit{Human-Only}) and using the \textit{Frontier} or \textit{Coverage} planners leveraging dense human labels.

\begin{figure}[!t]
    \centering
    \includegraphics[width=\columnwidth]{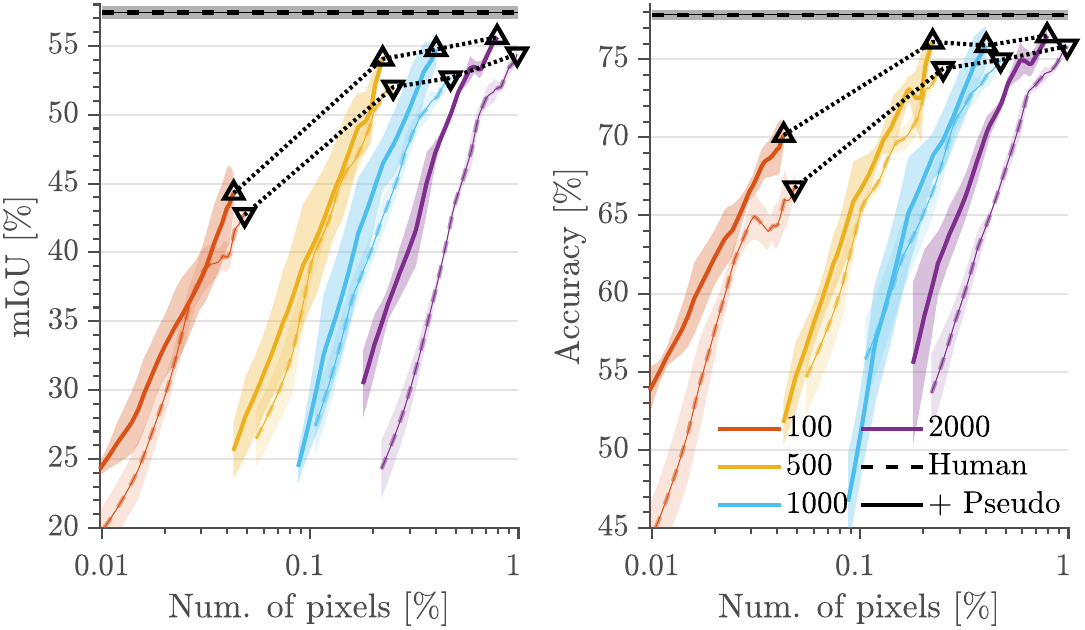}
    \caption{Comparison of our human label selection only (dashed transparent lines), and combined with our pseudo label selection (solid lines) over varying labelling budgets $\alpha \in [100, 2000]\,\text{px}$ per image using our frontier planner on ISPRS Potsdam. Results are averaged over three runs. Shaded regions indicate one standard deviation. Using our pseudo labels consistently improves performance.}
    \label{F:pseudo_human_labels_combined_varying_budget}
\end{figure}

\begin{figure}[!t]
    \captionsetup[subfigure]{labelformat=empty}

    \centering
    \subfloat[]{\includegraphics[width=0.19\columnwidth]{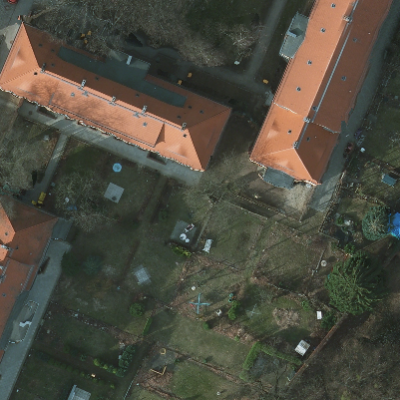}}
    \hfill
    \subfloat[]{\includegraphics[width=0.19\columnwidth]{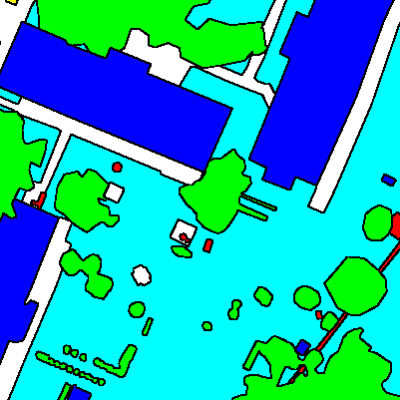}}
    \hfill
    \subfloat[]{\includegraphics[width=0.19\columnwidth]{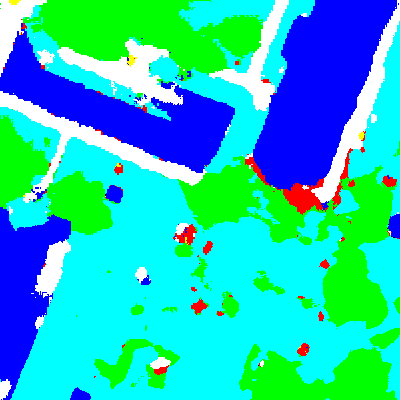}}
    \hfill
    \subfloat[]{\includegraphics[width=0.19\columnwidth]{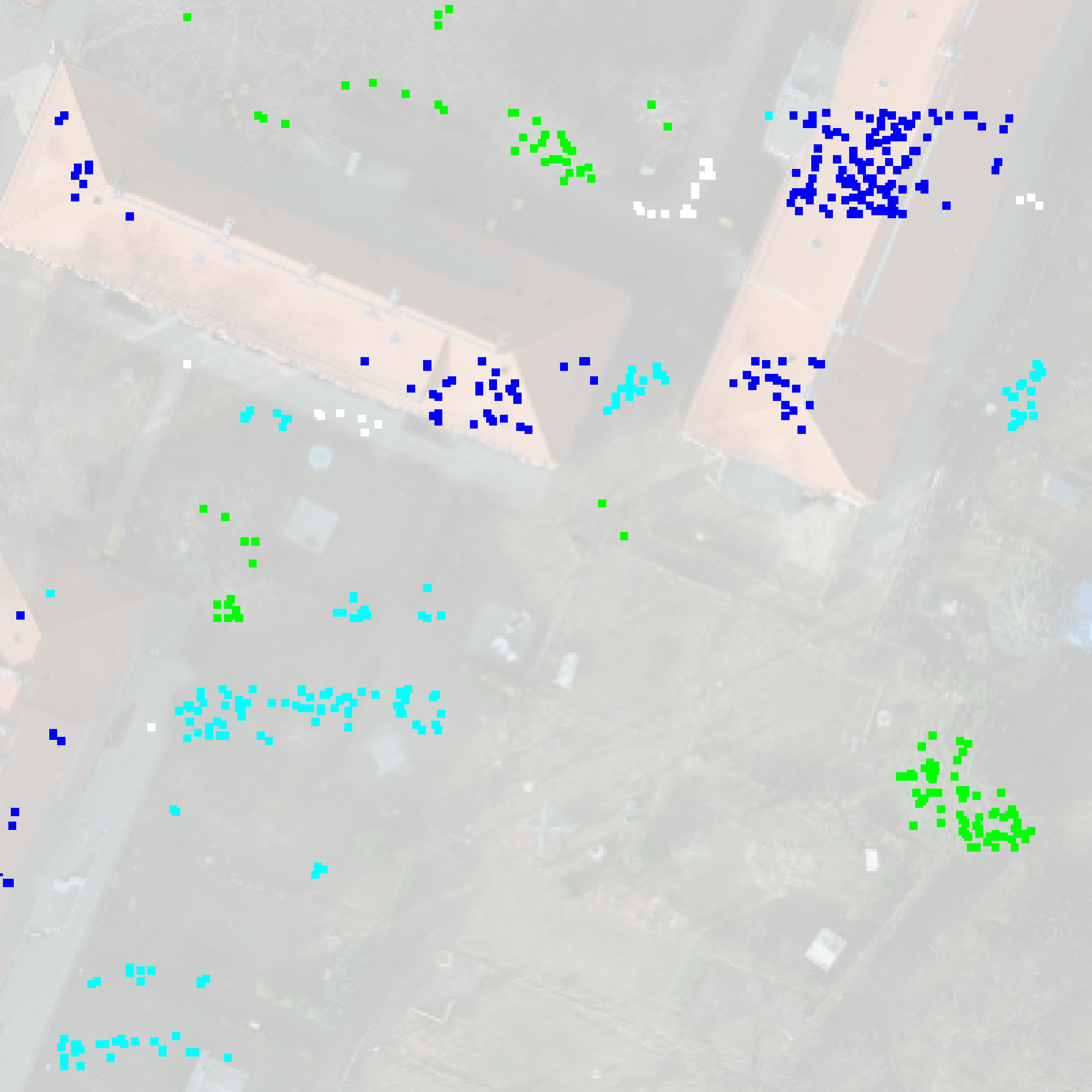}}
    \hfill
    \subfloat[]{\includegraphics[width=0.19\columnwidth]{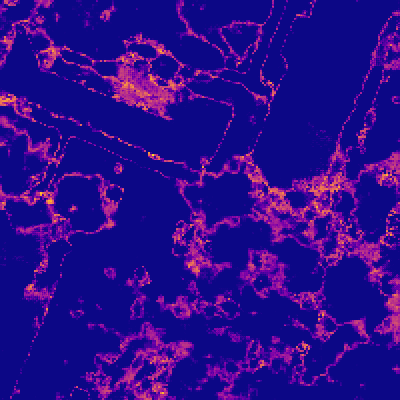}}

    \vspace{-4mm}
    \subfloat[]{\includegraphics[width=0.19\columnwidth]{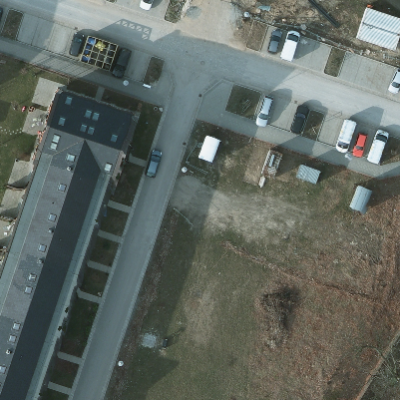}}
    \hfill
    \subfloat[]{\includegraphics[width=0.19\columnwidth]{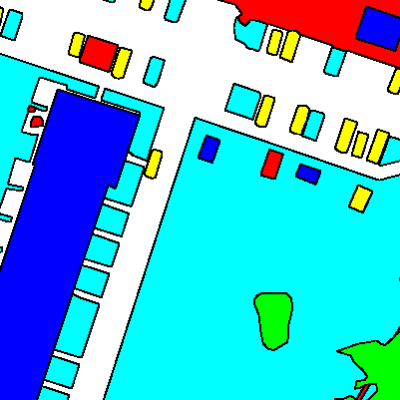}}
    \hfill
    \subfloat[]{\includegraphics[width=0.19\columnwidth]{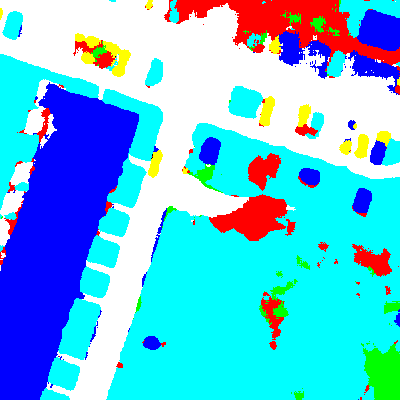}}
    \hfill
    \subfloat[]{\includegraphics[width=0.19\columnwidth]{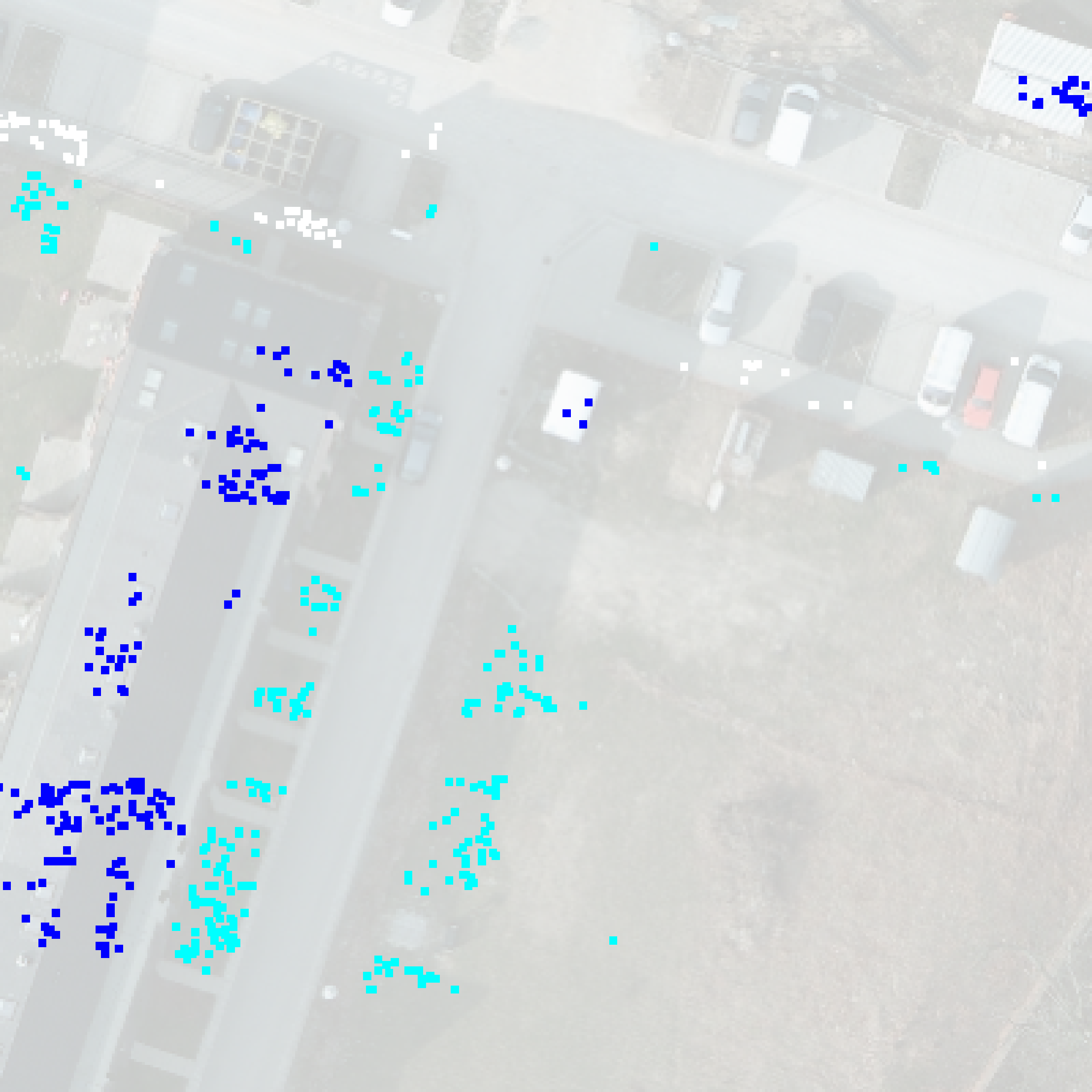}}
    \hfill
    \subfloat[]{\includegraphics[width=0.19\columnwidth]{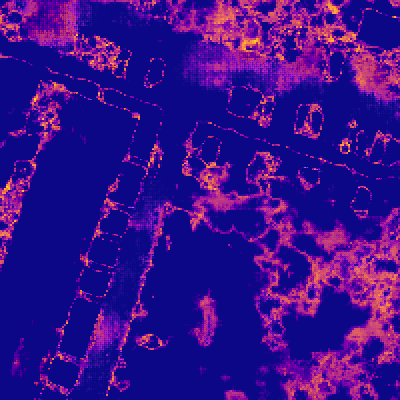}}

    \vspace{-4mm}
    \subfloat[Input]{\includegraphics[width=0.19\columnwidth]{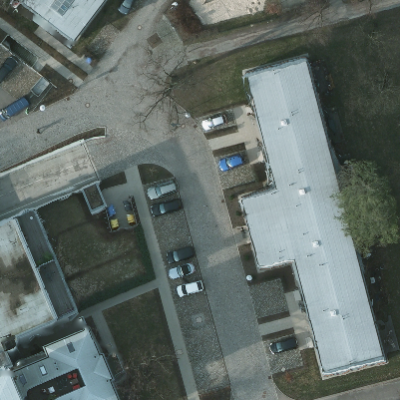}}
    \hfill
    \subfloat[Ground Truth]{\includegraphics[width=0.19\columnwidth]{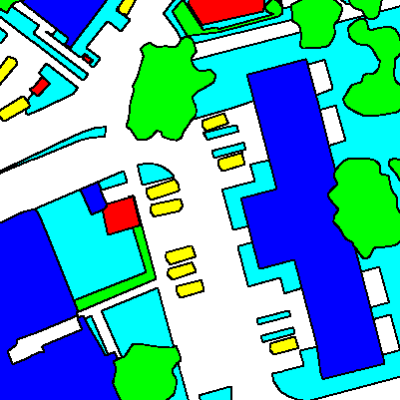}}
    \hfill
    \subfloat[Pseudo Label]{\includegraphics[width=0.19\columnwidth]{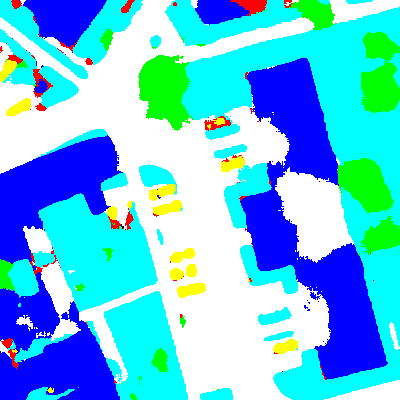}}
    \hfill
    \subfloat[Sparse Label]{\includegraphics[width=0.19\columnwidth]{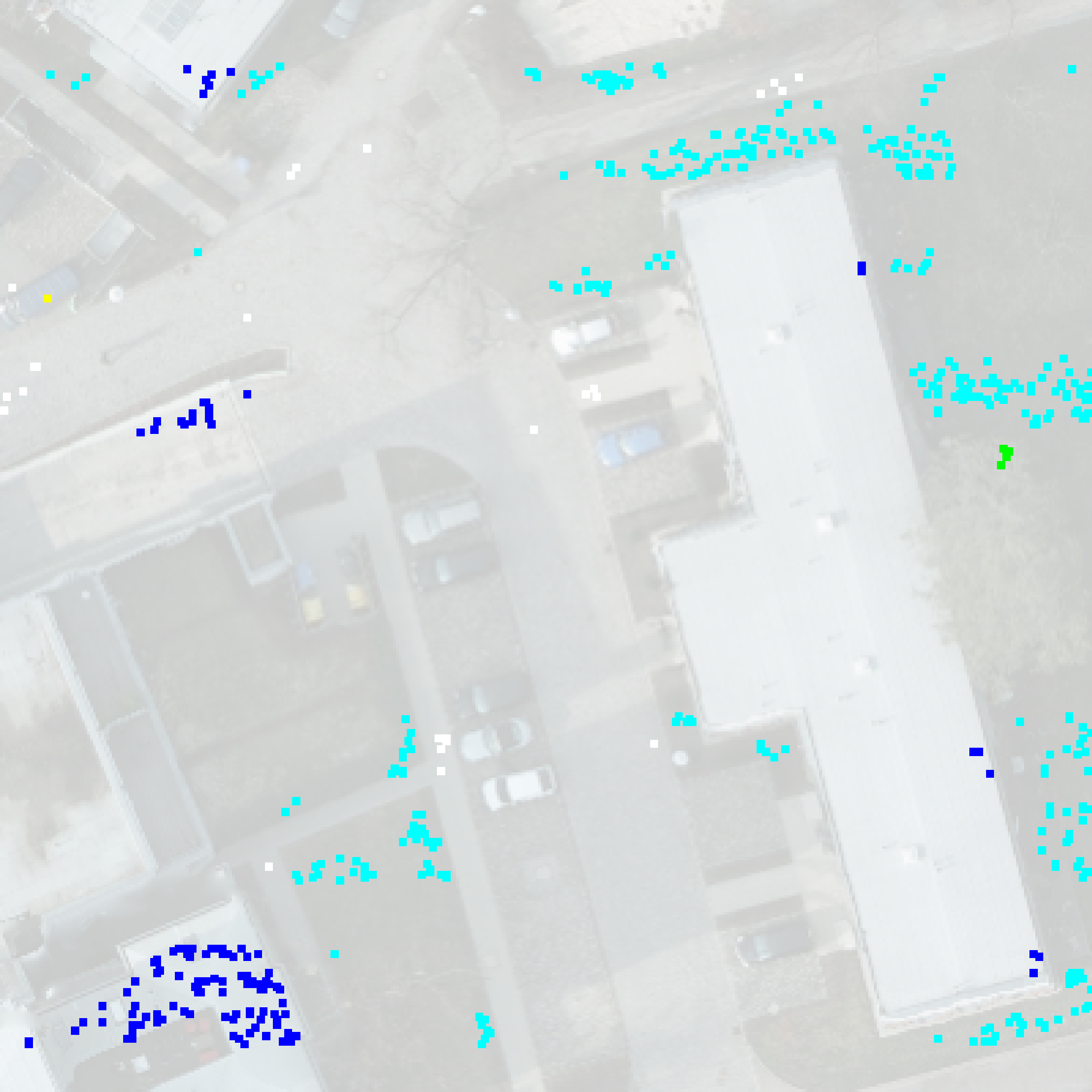}}
    \hfill
    \subfloat[Uncertainty]{\includegraphics[width=0.19\columnwidth]{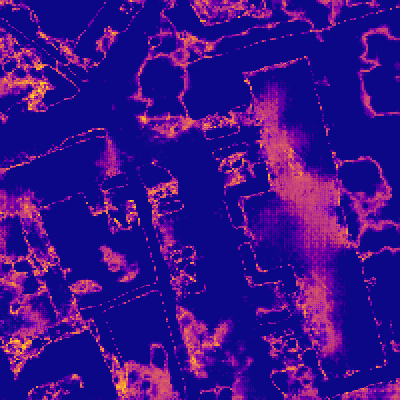}}

\caption{Qualitative results of our pseudo label generation on ISPRS Potsdam. Left to right: RGB input, ground truth, pseudo label, pixels selected for re-training, model uncertainty. Selected pixels are expanded to their one-pixel neighbourhood for visualisation. Our method selects low-uncertainty pixels to minimise label errors.}
\label{F:sparse_pseudo_labels_qualitative_results}
\end{figure}

\cref{F:pseudo_human_labels_combined_1000_pixels} summarises the performance of the different methods. Combining human labels with pseudo labels improves performance over sparse human labels only (green). This verifies our concept of leveraging both sparse human supervision and self-supervised pseudo labels to maximise performance. Our uncertainty-aware pseudo label selection method (dark blue) achieves $\approx 1-2\%$ \ac{mIoU} higher than other methods (purple, red). Particularly, our semi-supervised approach outperforms the \textit{Coverage} planner using only $\approx 0.6\,\%$ of the human-labelled pixels. \cref{F:sparse_pseudo_labels_qualitative_results} shows qualitative results for our method after mission completion.

\cref{F:pseudo_human_labels_combined_varying_budget} shows the performance using our human-labelled pixel selection method only (dashed transparent lines) and combining it with our uncertainty-aware pseudo labels (solid lines) over varying human labelling budgets $\alpha \in [0.06, 1.25]\,\%$ per image using our frontier planner. Combining sparse human and pseudo labels consistently improves performance by $\approx 2-3\,\%$ \ac{mIoU} across varying budgets. This validates the superior performance of our semi-supervised approach over using sparse human labels only. Further, our semi-supervised approach rapidly closes the final performance gap to the fully supervised frontier planner proposed by \citet{ruckin2023informative} (dashed black line). The fully supervised approach reaches a maximum performance of $\approx 57.5\%$ \ac{mIoU} while our semi-supervised approach reaches $\approx 56\%$ \ac{mIoU} with only $\approx 0.6\%$ of the human-labelled pixels. This shows that our semi-supervised approach requires two magnitudes fewer human-labelled pixels while reaching performance similar to previous fully supervised approaches.

\subsection{Semi- vs. Self-Supervised Robotic Active Learning} \label{SS:exp_ssl_for_al} 

\begin{figure}[!t]
    \centering
    \includegraphics[width=\columnwidth]{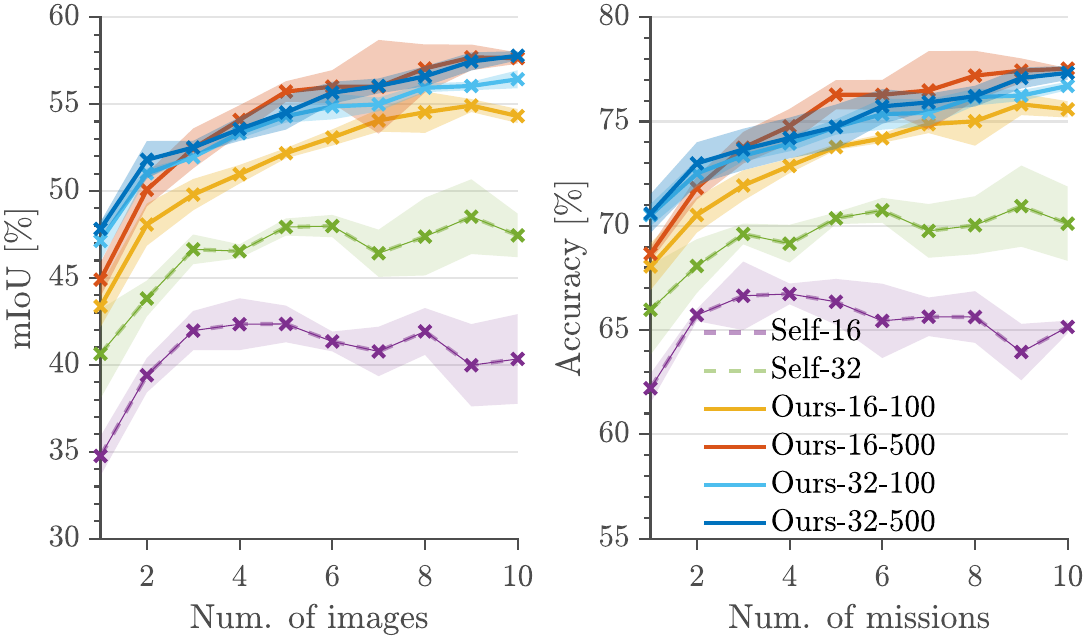}
    \caption{Comparison of our semi-supervised (solid lines) and a self-supervised approach (dashed transparent lines) with varying numbers of human-labelled pixels during deployment and densely human-labelled images for pre-training. Results are averaged over three runs on ISPRS Potsdam. Shaded regions indicate one standard deviation. Our semi-supervised approach outperforms the self-supervised approach for all labelling budget configurations.}
    \label{F:semi_vs_self_supervised_potsdam}
\end{figure}

The third set of experiments shows that our semi-supervised active learning framework outperforms self-supervised approaches by a large margin under varying human labelling budgets for model pre-training and model re-training in the unknown environment.

Similar to self-supervised approaches for robotic continual learning and domain adaptation~\citep{frey2021continual, zurbrugg2022embodied}, we utilise our frontier planner to guide uncertainty-driven training data collection and exploit the online-built map to generate dense pseudo labels. Current self-supervised approaches only work with pre-trained semantic segmentation models deployed in similar environments~\citep{frey2021continual, zurbrugg2022embodied, chaplot2021seal}. Although our semi-supervised method works in a completely unknown environment (\Cref{SS:exp_pseudo_label_generation}), for comparing to self-supervised methods, we relax these assumptions and consider small amounts of densely human-labelled pre-training data randomly sampled from the deployment environment. Each approach starts with the same model checkpoint trained on the sampled pre-training data. Similar to the experience replay method of self-supervised approaches~\citep{frey2021continual, zurbrugg2022embodied}, to achieve performance improvements in the self-supervised approach, the human-labelled pre-training data is additionally used for model re-training after a mission is completed.

\cref{F:semi_vs_self_supervised_potsdam} shows the semantic segmentation performance of our semi-supervised approach (solid lines) compared to the self-supervised approach (dashed lines) on ISPRS Potsdam with varying numbers of human-labelled pre-training data $\{16, 32\}$. For all human labelling budgets $\alpha \in \{100, 500\} \approx \{0.06, 0.3\}\%$ and all pre-training data budgets, our semi-supervised active learning approach outperforms self-supervision by a large margin. With a small number of $16$ pre-training images and little human supervision of $\alpha = 100$ during the missions, our semi-supervised approach achieves higher final performance than the self-supervised approach with $32$ pre-training images. Further, irrespective of the number of pre-training images, self-supervision fails to improve its performance after five missions. This suggests that semi-supervised active learning is necessary for maximally improving semantic segmentation in unknown or partially known environments. Although self-supervision benefits from minimal labelling requirements during deployment, it is inherently limited by its lack of knowledge and systematic prediction errors in unknown environments~\citep{chaplot2021seal}.

\section{Conclusions and Future Work} \label{S:conclusions}

We proposed a novel adaptive informative path planning approach for semi-supervised active learning in robotic semantic perception with minimal human labelling effort. Our main contribution is a method for selecting sparse sets of informative pixels for human labelling and combining them with automatically generated pseudo labels rendered from an online-built uncertainty-aware semantic map. Our experimental results show that our sparse human-labelled pixel selection method outperforms state-of-the-art pixel selection methods. Combining human labels with pseudo labels further improves performance. Our semi-supervised approach drastically reduces human labelling effort compared to fully supervised methods while preserving similar performance and outperforms purely self-supervised approaches. Despite those encouraging results, future work could develop new methods to generate human labelling queries, e.g. by utilising foundation models for image segmentation~\citep{Kirillov2023sam}, and automatically extract pseudo labels. Our framework could foster this new research in robotic active learning.

\bibliographystyle{IEEEtranN}
\footnotesize
\bibliography{2024-icra-rueckin}

\end{document}

%% file: additionals/common_acronyms.tex
\begin{acronym}

\acro{2D}{two-dimensional}

\acro{3D}{three-dimensional}

\acro{AHRS}{attitude and heading reference system}

\acro{AUV}{autonomous underwater vehicle}

\acro{CPP}{Chinese Postman Problem}

\acro{DoF}{degree of freedom}
\acrodefplural{DoF}[DoFs]{degrees of freedom}

\acro{DVL}{Doppler velocity log}

\acro{FSM}{finite state machine}

\acro{IMU}{inertial measurement unit}

\acro{LBL}{Long Baseline}

\acro{MCM}{mine countermeasures}

\acro{MDP}{Markov decision process}
\acrodefplural{MDP}[MDPs]{Markov decision processes}

\acro{POMDP}{Partially Observable Markov Decision Process}
\acrodefplural{POMDP}[POMDPs]{Partially Observable Markov Decision Processes}

\acro{PRM}{Probabilistic Roadmap}
\acrodefplural{PRM}[PRM]{Probabilistic Roadmaps}

\acro{ROI}{region of interest}
\acrodefplural{ROI}[ROIs]{regions of interest}

\acro{ROS}{Robot Operating System}

\acro{ROV}{remotely operated vehicle}

\acro{RRT}{Rapidly-exploring Random Tree}
\acrodefplural{RRT}[RRTs]{Rapidly-exploring Random Trees}

\acro{SLAM}{Simultaneous Localisation and Mapping}

\acro{SSE}{sum of squared errors}

\acro{STOMP}{Stochastic Trajectory Optimization for Motion Planning}

\acro{TRN}{Terrain-Relative Navigation}

\acro{UAV}{unmanned aerial vehicle}

\acro{USBL}{Ultra-Short Baseline}

\acro{IPP}{informative path planning}

\acro{FoV}{field of view}
\acrodefplural{FoV}[FoVs]{fields of view}

\acro{CDF}{cumulative distribution function}

\acro{ML}{maximum likelihood}

\acro{RMSE}{Root Mean Squared Error}
\acro{MLL}{Mean Log Loss}

\acro{GP}{Gaussian Process}
\acrodefplural{GP}[GPs]{Gaussian Processes}

\acro{KF}{Kalman Filter}

\acro{IP}{Interior Point}
\acro{BO}{Bayesian Optimization}

\acro{SE}{squared exponential}

\acro{UI}{uncertain input}

\acro{MCL}{Monte Carlo Localisation}
\acro{AMCL}{Adaptive Monte Carlo Localisation}

\acro{SSIM}{Structural Similarity Index}

\acro{MAE}{Mean Absolute Error}
\acro{RMSE}{Root Mean Squared Error}
\acro{AUSE}{Area Under the Sparsification Error curve}

\acro{AL}{active learning}
\acro{DL}{deep learning}
\acro{CNN}{convolutional neural network}

\acro{MC}{Monte-Carlo}

\acro{GSD}{ground sample distance}

\acro{BALD}{Bayesian active learning by disagreement}

\acro{fCNN}{fully convolutional neural network}
\acro{FCN}{fully convolutional neural network}

\acro{CMA-ES}{covariance matrix adaptation evolution strategy}

\acro{FoV}{field of view}

\acro{mIoU}{mean Intersection-over-Union}
\acro{ECE}{expected calibration error}

\acro{GAN}{Generative Adversarial Network}
\acro{POMCP}{Partially Observable Monte-Carlo Planning}
\acro{MCTS}{Monte-Carlo tree search}
\acro{RL}{reinforcement learning}

\end{acronym}